\pdfoutput=1

\documentclass[11pt]{article}

\usepackage{acl}

\usepackage{times}
\usepackage{latexsym}
\usepackage{multirow}
\usepackage{booktabs}
\usepackage{xcolor}
\usepackage{amsmath}
\usepackage{mathabx}
\usepackage[T1]{fontenc}

\usepackage[utf8]{inputenc}

\usepackage{microtype}

\usepackage{inconsolata}

\usepackage{graphicx}
\usepackage{tabularx}
\usepackage{colortbl}
\usepackage{tcolorbox}
\title{Enhancing Logical Reasoning in Language Models via Symbolically-Guided Monte Carlo Process Supervision}

\author{Xingwei Tan$^{1}$, Marco Valentino$^{1}$, Mahmud Akhter$^{2}$, Maria Liakata$^{2,3}$, Nikolaos Aletras$^{1}$ \\
    $^1$School of Computer Science, University of Sheffield\\
    $^2$School of Electronic Engineering and Computer Science, Queen Mary University of London\\
    $^3$The Alan Turing Institute\\
    \texttt{\{xingwei.tan,m.valentino,n.aletras\}@sheffield.ac.uk}\\
    \texttt{\{m.akhter,m.liakata\}@qmul.ac.uk}}

\begin{document}
\maketitle
\begin{abstract}
Large language models (LLMs) have shown strong performance in many reasoning benchmarks.
However, recent studies have pointed to memorization, rather than generalization, as one of the leading causes for such performance.
LLMs, in fact, are susceptible to content variations, demonstrating a lack of robust planning or symbolic abstractions supporting their reasoning process. 
To improve reliability, many attempts have been made to combine LLMs with symbolic methods.
Nevertheless, existing approaches fail to effectively leverage symbolic representations due to the challenges involved in developing reliable and scalable verification mechanisms.
In this paper, we propose to overcome such limitations by synthesizing high-quality symbolic reasoning trajectories with stepwise pseudo-labels at scale via Monte Carlo estimation.
A Process Reward Model (PRM) can be efficiently trained based on the synthesized data and then used to select more symbolic trajectories.
The trajectories are then employed with Direct Preference Optimization (DPO) and Supervised Fine-Tuning (SFT) to improve logical reasoning and generalization.
Our results on benchmarks (i.e., FOLIO and LogicAsker) show the effectiveness of the proposed method with gains on frontier and open-weight models.
Moreover, additional experiments on claim verification data reveal that fine-tuning on the generated symbolic reasoning trajectories enhances out-of-domain generalizability, suggesting the potential impact of the proposed method in enhancing planning and logical reasoning.\footnote{Our experimental code and data: \url{https://github.com/Xingwei-Tan/Symbolic-Guided_MC}}

\end{abstract}

\section{Introduction}

Large language models (LLMs) have demonstrated strong capabilities across a variety of NLP tasks in different domains \cite{srivastava2023beyond}.
As language is the primary medium through which humans formulate logical arguments, researchers have focused on exploring whether logical reasoning capabilities can emerge in LLMs \cite{NEURIPS2022_8bb0d291}.
Recent studies have indeed found evidence of emergent reasoning capabilities, where LLMs, with proper guidance, can mimic humans' multi-step thinking process \cite{wei2022chain,NEURIPS2022_8bb0d291,yao2023react}.

\begin{figure}[t]
    \centering
    \includegraphics[width=\linewidth]{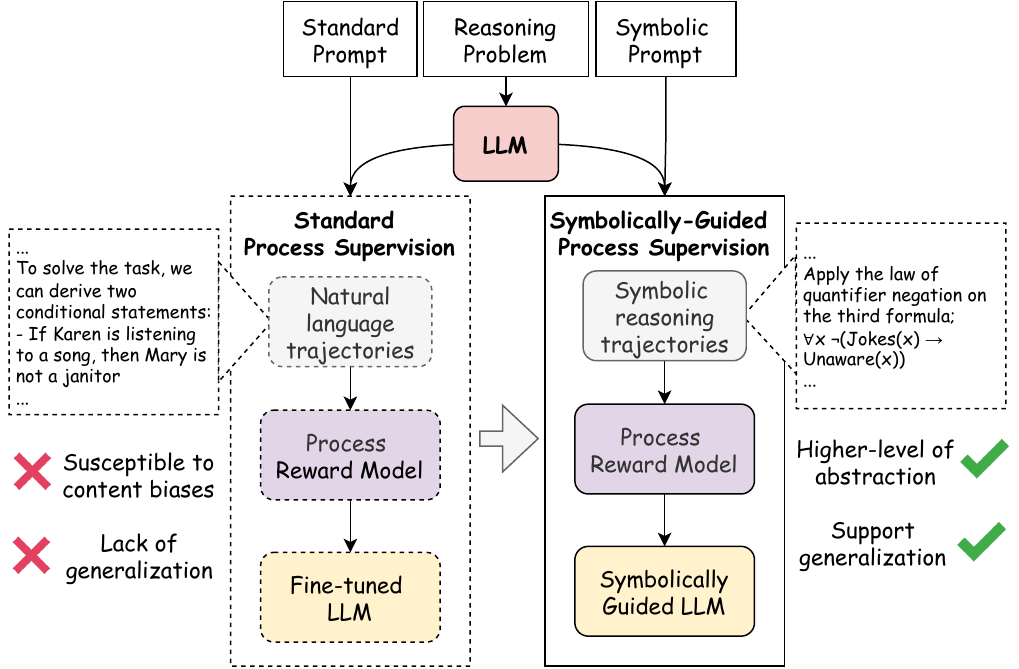}
    \caption{Standard offline simulation methods for reasoning rely on reasoning trajectories entirely expressed in natural language. For logical reasoning tasks, however, natural language trajectories are prone to content biases, leading to a lack of generalization. We present a novel symbolically-guided process synthesizing method to derive LLMs with enhanced logical capabilities through high-quality symbolic reasoning trajectories.}
    \label{fig:intro}
\end{figure}

While chain-of-thought reasoning might appear plausible at the surface, subsequent studies have found it unfaithful and contribute little to the final answer \cite{sprague2025to,lewislim2025analysingchainthoughtdynamics}. 
When multiple valid deductive steps are available, for instance, LLMs are incapable of planning to systematically explore different possibilities \cite{saparov2023language}.
To address this limitation, prompting strategies have been combined with search methods. For example, \citet{yao2023tree} propose Tree of Thoughts to search over actions and states, mimicking \textit{``slow thinking''}, which allows exploring as many available directions as possible before providing the final answer.
Similarly, \citet{hao-etal-2023-reasoning} frame the reasoning process as a Markov Decision Process, where LLMs generate actions and world observations to simulate planning.

However, performing an extensive search to plan reasoning at inference time (i.e., \textit{online simulation}) inevitably results in high latency \cite{jiao-etal-2024-learning}.
On the other hand, \textit{offline simulation} \cite{jiao-etal-2024-learning} samples reasoning trajectories and uses Monte Carlo estimation to evaluate the quality of the steps.
The trajectories and their pseudo-labels are then used to train a PRM, which is later used to rank and select more trajectories for fine-tuning to integrate the planning-based reasoning into the LLMs.
However, this method is still limited to trajectories entirely expressed in natural language. 
This reliance on concrete logical arguments makes the models susceptible to content and prompt variations \cite{xu-etal-2024-faithful}, often limiting their ability to generalize beyond the training domain \cite{dougrezlewis2025assessingreasoningcapabilitiesllms}.

Formal languages have been proposed as explicit representations to enhance faithfulness and robustness in logical reasoning, thereby facilitating the integration of Large Language Models (LLMs) with symbolic methods \cite{pan-etal-2023-logic}. 
Unfortunately, existing approaches struggle to effectively leverage symbolic representations. 
This is primarily, due to the inherent challenges in developing scalable verification mechanisms, which are hampered by the rigidity and complexity of external symbolic solvers \cite{ranaldi2025improvingchainofthoughtreasoningquasisymbolic}, and in establishing reliable quality checks, given the systematic nature of logical reasoning \cite{quan-etal-2024-verification}.

\begin{figure*}[!tb]
    \centering
    \includegraphics[width=\linewidth]{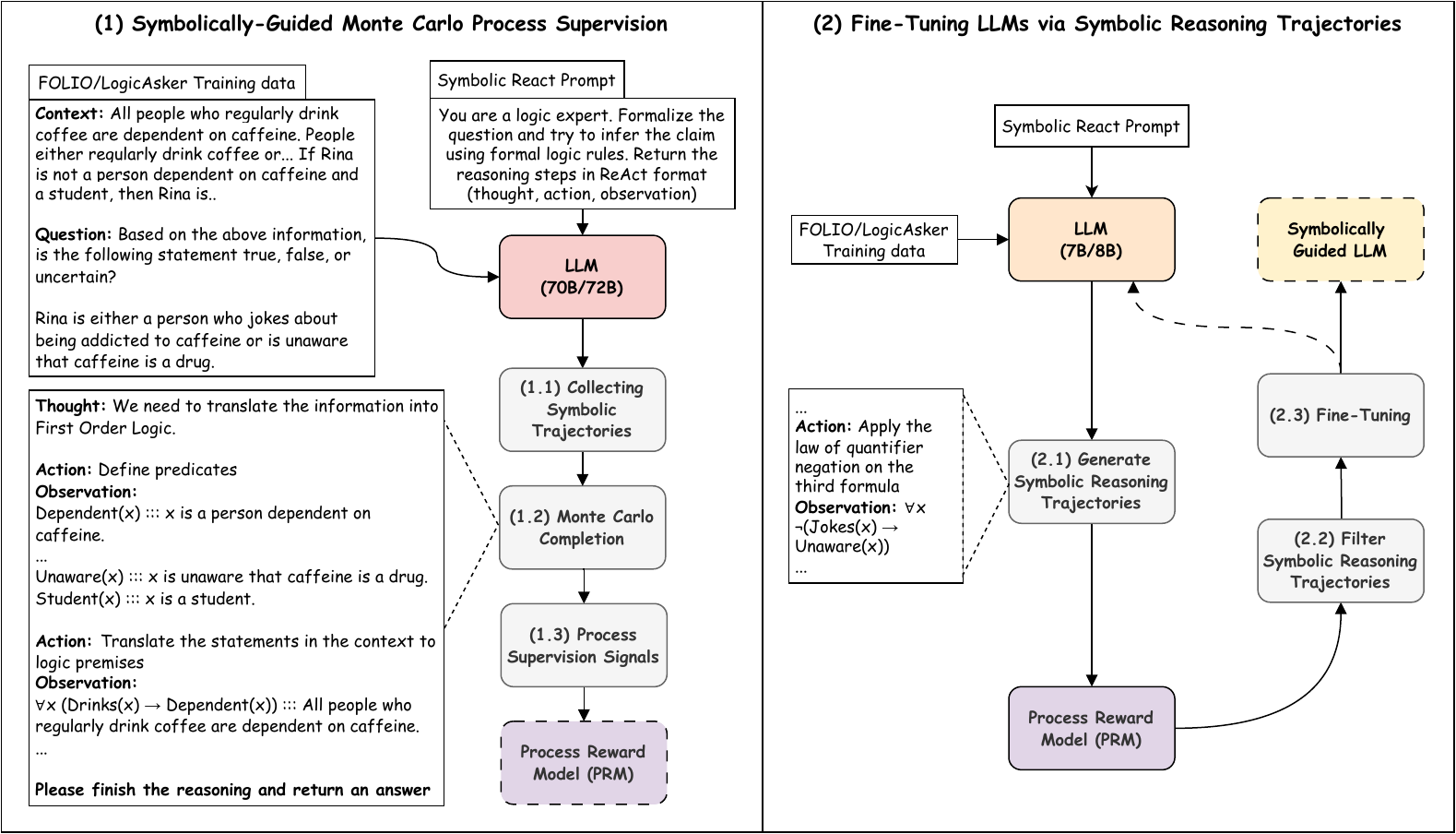}
    \caption{Overall pipeline for enhancing LLM with synthesized symbolic trajectories produced from Monte Carlo process supervision. 1) We propose an extension to the ReAct method (i.e. Symbolic ReAct) to collect symbolic reasoning trajectories and employ Monte Carlo process supervision to train a Process Reward Model (PRM). Subsequently, the PRM is adopted in interaction with smaller models to support fine-tuning on high-quality symbolic reasoning trajectories.}
    \label{fig:pipeline}
\end{figure*}

In this paper, we propose to address the limitations in offline planning-based reasoning simulation and symbolic-based LLM reasoning by proposing a new method to synthesize high-quality symbolic reasoning trajectories (see Figure~\ref{fig:intro}). Then combine it with Monte Carlo estimation \cite{jiao-etal-2024-learning,wang-etal-2024-math} to produce symbolic reasoning trajectories with stepwise pseudo-labels at scale to support subsequent offline training for enhancing logical reasoning (see Figure~\ref{fig:pipeline}).
Specifically, our methodology is centred around the development of Process Reward Models (PRMs) \cite{lightman2024lets} to automatically determine the quality of symbolic reasoning trajectories for logical reasoning. To this end, we first propose an extension to ReAct \cite{yao2023react} (i.e., Symbolic ReAct) to collect symbolic reasoning trajectories. Subsequently, we employ Monte Carlo estimation to produce stepwise signals based on larger LLMs (i.e., 70/72B parameters) to train a PRM. Finally, the PRM is adopted in interaction with more trajectories produced by smaller LLMs (i.e., 7/8B parameters) to support fine-tuning methods such as SFT \cite{10.5555/3600270.3602281} and DPO \cite{rafailov2023direct} without additional human supervision to derive models that can perform logical reasoning by leveraging explicit symbolic representations.

Extensive experiments on logical reasoning benchmarks: FOLIO \cite{han-etal-2024-folio} and LogicAsker \cite{wan-etal-2024-logicasker} demonstrate the effectiveness of our proposed method. 
In particular, we found that symbolic ReAct can improve the performance of frontier models when adopted as a prompting strategy (i.e., +6\% for GTP-5~\cite{openai2025gpt5card} and +26\% for Deepseek-V3~\cite{deepseekai2025deepseekv3technicalreport} over SymbCoT~\cite{xu-etal-2024-faithful} and +4\% and +7\% over non-symbolic ReAct), and contribute to the development of effective PRMs for enhancing smaller language models (i.e., +9\% Llama 8b and +4\% Qwen 7b).

Moreover, to investigate how the fine-tuned models reason on real-world tasks, we perform out-of-domain evaluation on claim verification datasets \cite{dougrezlewis2025assessingreasoningcapabilitiesllms} to test their generalizability. The results reveal that fine-tuning LLMs with the proposed framework enhances the performance on claim verification compared to the baseline, suggesting the proposed method produces more robust models.

We release the generated symbolic reasoning trajectories with stepwise pseudo-labels as part of a novel dataset, \textit{SymbReAct-trace}\footnote{\url{https://huggingface.co/collections/XingweiT/symbreact-trace-68c422ec5edf4750d2f01add}}, to support future research at the intersection of logical reasoning and generalization with LLMs.

\section{Related Work}

\subsection{Enhancing LLM Reasoning}

Recent work has explored ways to enhance the reasoning abilities of LLMs.
Early work used textual instructions to prompt and then generate reasoning steps before reaching a final conclusion \cite{10.5555/3600270.3602070,NEURIPS2022_8bb0d291,yao2023react}.
Later work tried to incorporate searching, which encourages LLMs to explore multiple reasoning directions.
\citet{yao2023tree} break reasoning into steps and form a tree at each step, then find the best reasoning trace by calculating the fast rewards with breadth-first search or depth-first search. 
\citet{hao-etal-2023-reasoning} instead model the step-by-step reasoning as a Markov Decision Process and apply Monte Carlo Tree Search to search for the optimal reasoning plan.
Although these methods can be directly plugged to off-the-shelf LLMs, they require an extensive amount of sampling at inference and thus are impractical applications that require low latency.

Another line of work tries to enhance LLM reasoning via fine-tuning or preference optimization.
\citet{zelikman2022star} propose to bootstrap LLMs by guiding them in generating reasoning traces, then select those that could reach the correct answers.
However, wrong reasoning traces can also lead to correct answers, and learning from them can hurt a model's ability to solve more challenging problems.
\citet{lightman2024lets} show that a process reward model trained on large-scale data with step-wise signals can better predict the final answer than an outcome reward model.
Such a process reward model can help select a reasoning trace so it not only has a correct outcome but also ensures reliability of intermediate steps.
\citet{luo2024improvemathematicalreasoninglanguage} propose an automatic method to train process reward models on a divide-and-conquer style Monte Carlo tree search, avoiding expensive human annotations.
\citet{wang-etal-2024-math} introduce Monte Carlo estimation for generating stepwise pseudo-labels, which treats the quality of a reasoning step as its potential to deduce the correct answer and estimate the step quality by generating multiple completions.
\citet{jiao-etal-2024-learning} propose offline simulation for learning planning-based reasoning via synthesizing reasoning steps in ReAct~\cite{yao2023react} format and estimate the stepwise correctness with Monte Carlo completions.
However, these methods do not consider the thinking styles of the trajectories, which is essential to solving complex logical questions.

\subsection{Symbolic Reasoning with LLMs}

A different line of work combines the reliability of symbolic systems and the flexibility of LLMs.
One approach is to fine-tune LLMs on the formal proving steps generated by proof assistant, such as Lean, to improve their performance on theorem-proving problems \cite{xin2024deepseekproveradvancingtheoremproving,xin2025deepseekproverv}.
Trying to tackle a wider range of real-world problems, another line of work proposes using LLMs to automatically formalize natural language questions into symbolic representations (auto-formalization), then use an external solver to infer a conclusion \cite{pan-etal-2023-logic,quan-etal-2024-verification,wysocka-etal-2025-syllobio}.
However, these methods face the problem that auto-formalization often fails to produce valid symbolic representations for real-world questions.
To improve flexibility, \citet{ranaldi2025improvingchainofthoughtreasoningquasisymbolic} propose to guide LLMs to operate at a high level of abstraction of the questions via quasi-symbolic explanations.
On the other hand, \citet{xu-etal-2024-faithful} use off-the-shelf LLMs as the solver, which can tolerate a certain degree of errors in the symbolic representations.
The above-mentioned work all focus on logical benchmarks in abstractive settings, while our work offers improvements in both standard logical reasoning benchmarks and in real-world problems.

\section{Symbolically-Guided Monte Carlo Process Supervision}

%

Since LLMs are mostly trained on natural language data, without explicit guidance, they tend to formulate reasoning trajectories without leveraging symbolic abstractions or formal representations.
While arguments expressed in natural language are viable for addressing common reasoning problems, recent studies have shown that they possess limited efficacy, robustness, and faithfulness when it comes to systematic logical reasoning \cite{10.1162/tacl_a_00594,lyu-etal-2023-faithful,turpin2023language,yee2024faithful}.

We aim to steer LLMs via prompting to generate reasoning trajectories that involve explicit symbolic formalisms. To this end, we propose an extension to ReAct designed to elicit symbolic reasoning trajectories, which we call \emph{Symbolic ReAct} (\ref{sec:sampling}).
Then, we perform Monte Carlo estimation to assess the quality of intermediate reasoning steps (\ref{sec:prm}).
The pseudo-labels generated through Monte Carlo estimation are used to train a Process Reward Model \citep[PRM]{lightman2024lets}.
Finally, we leverage the resulting high-quality trajectories selected based on the PRM to fine-tune smaller LLMs \ref{sec:data}. 
Figure \ref{fig:pipeline} shows the overall process.

\subsection{Collecting Reasoning Trajectories via Symbolic ReAct}

Given a logical reasoning dataset \(\mathcal{D} = \{(P_i, H_i, y_i)\}_{i=1}^N\), where each instance consists of a set of premises \(P_i\), a hypothesis \(H_i\), and a label \(y_i \in \{\texttt{True}, \texttt{False}\}\) indicating whether \(H_i\) logically follows from \(P_i\), we construct symbolic reasoning trajectories in a \emph{Symbolic ReAct} format. Each trajectory comprises a sequence of steps structured as \textit{thoughts}, \textit{actions}, and \textit{observations} (see Figure~\ref{fig:pipeline}, left).

For a given problem instance \((P, H)\), the goal is to determine whether \(P \models H\). Here, a \textit{thought} \(t_j\) represents a high-level reasoning step or plan toward solving the problem. Each thought \(t_j\) may be followed by one or more \textit{actions} \(a_{j,k}\), which are executable symbolic operations (e.g., formalization, rule application) designed to advance the reasoning state. Executing an action results in an updated environment state, captured as an \textit{observation} \(o_{j,k} = \texttt{Execute}(a_{j,k})\), reflecting the new knowledge or inference derived from the action. A reasoning trajectory can be represented as:
\[
(t_1, a_{1,1}, o_{1,1}, a_{1,2}, o_{1,2}, \ldots, t_2, a_{2,1}, o_{2,1}, \ldots)
\]
where each thought \(t_j\) leads to a variable-length sequence of \((a_{j,k}, o_{j,k})\) pairs.

Symbolic ReAct trajectories are generated via in-context learning, using one-shot prompting to illustrate the desired structure and reasoning process (see the prompt Appendix \ref{sec:appendix3} Table \ref{symbolic_prompt_table}). The examples are formalized in \textit{first-order logic (FOL)} to encode thoughts, actions, and observations symbolically, thereby enabling precise logical manipulation and traceability.

Our investigation focuses on two key questions: 
\vspace{-8pt}
\begin{enumerate}
    \item Can symbolic guidance via ReAct improve LLMs' logical reasoning capabilities?
    \vspace{-8pt}
    \item Can the structured symbolic trajectories serve as effective supervision signals for fine-tuning techniques?
\end{enumerate}

\label{sec:sampling}

\definecolor{rowcolor}{gray}{0.9}
\definecolor{linecolor}{gray}{0.6}

\subsection{Monte Carlo Process Supervision}
\label{sec:prm}

Sound and robust logical reasoning crucially depends on the ability to derive formally valid conclusions through the correct application of valid reasoning schemas.
Therefore, to achieve high performance and generalization, LLMs should learn to apply valid reasoning patterns and logical schemes across different problems.
To this end, we define a process reward model to provide detailed step-wise feedback over the generated reasoning trajectories.
A process reward model is usually trained using the following loss function:
\begin{equation}
    \mathcal{L}_{\text{PRM}}=-\sum^{N}_{i=1}\Big[y_{s_i} \log \hat{y}_{s_i}+(1-y_{s_i}) \log (1-\hat{y}_{s_i})\Big],
    \label{eq:PRM_objective}
\end{equation}
where $y_{s_i}$ is the golden label of $s_i$ (the $i$-th step in the reasoning trace $s$), $\hat{y}_{s_i}$ is the score on $s_i$ predicted by the model, and $N$ is the total number of reasoning steps.

Ideally, a PRM can be trained on human-annotated reasoning steps where each step is verified based on its soundness and formal validity \cite{lightman2024lets}.
However, such an annotation process is not scalable in practice due to the complexity involved in formal verification.
Therefore, following recent work, we approximate stepwise correctness by using Monte Carlo estimation\cite{jiao-etal-2024-learning,wang-etal-2024-math}.

\begin{figure}[t]
    \centering
    \includegraphics[width=0.7\linewidth]{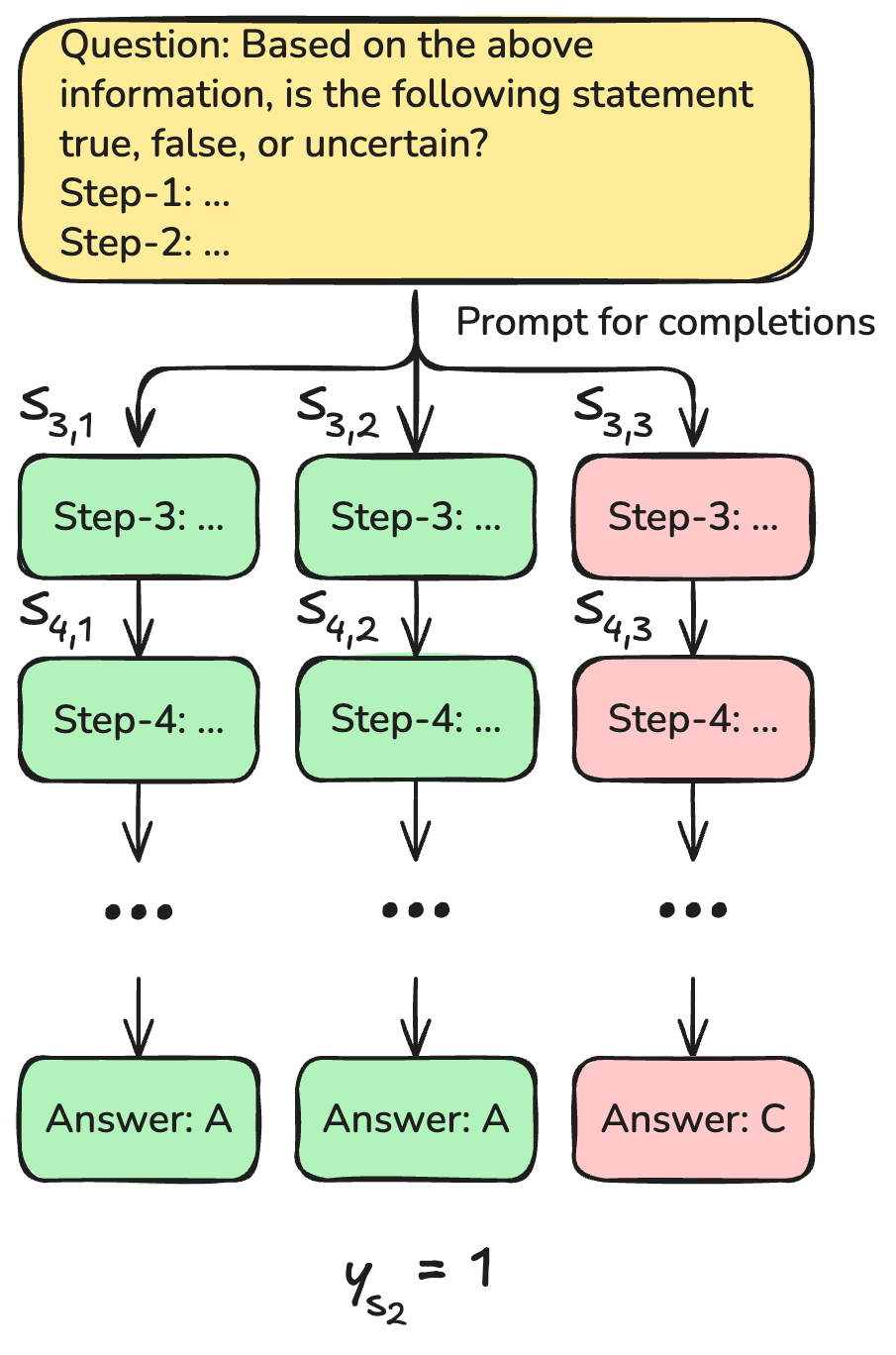}
    \caption{The concept of Monte Carlo estimation. Green blocks are completions reaching the correct answer.}
    \label{fig:mc_estimation}
\end{figure}

We perform Monte Carlo estimation (see Fig. \ref{fig:mc_estimation}) on top of the reasoning trajectories generated via the symbolic ReAct method described in Section \ref{sec:sampling}.
Specifically, we prompt an LLM to complete a partial solution generated via Symbolic ReAct and extracted from the collected reasoning trajectories -- i.e., providing the first $n$ steps in context and asking it to derive the final answer after completing the reasoning (see Appendix \ref{sec:appendix3} Table \ref{completion_prompt}).
We then verify whether the completions reach the correct answer, and assign a pseudo label for the intermediate step where the completions start \cite{zhang2025lessonsdevelopingprocessreward}.

In particular, after generating 10 samples of completion for each seed trajectory, the pseudo label is determined by matching the predicted answer with the ground truth to check how many completions are successful.
Following \citet{zhang2025lessonsdevelopingprocessreward}, we define hard binary labels (i.e., correct vs incorrect) as pseudo-labels since models trained on hard labels have been shown to achieve better performance than soft labels (i.e., based on a continuous score).
To determine the hard labels, we simply assign a positive score (i.e., +1) to a reasoning step if at least one of the sampled completions reaches the correct answer; otherwise, a negative score is assigned (i.e., -1).
The number of completions sampled for each trajectory and the number of successful completions needed for assigning a positive label are both hyperparameters. 
After acquiring the pseudo-labels for the reasoning trajectories, we use the collected data to fine-tune a PRM.
It is also worth noting that the effectiveness of Monte Carlo estimation heavily depends on the LLMs used for completions.
To minimize the chance that the LLM completers fail to reach a correct answer due to their own inability, we use the most capable open-weight LLMs we can host on our devices (i.e., \textit{Llama3.3-70B-Instruct} and \textit{Qwen2.5-72B-Instruct}).

\section{Synthetic Dataset Generation \& Fine-Tuning}
\label{sec:data}

We apply our symbolic trajectory collection and annotation process on two widely adopted logical reasoning datasets: FOLIO \cite{han-etal-2024-folio} and LogicAsker \cite{wan-etal-2024-logicasker}, which strike a balance between complexity and scalability.

\paragraph{FOLIO.} This is an expert-written, logically complex and diverse dataset for reasoning. Each sample consists of a context and a statement, which requires the model to determine whether the statement is true, false, or uncertain. It contains logic connectives such as implications, conjunctions, and disjunctions. The authors only provide a training split and a development split for FOLIO and FOLIOv2. As FOLIO has been used in many existing papers, we use the development set of it as the test set and the development set of FOLIOv2 as the development set for our experiment.

\paragraph{LogicAsker.} This dataset encompasses a comprehensive set of atomic logical rules, integrating them to construct queries that require long inference chains. It requires the model to determine whether the target statement is true or false, given the context. In our experiments, we generated $900$ question-answer pairs with the inference chain length set to $7$, $8$, or $9$. Then, they were split evenly into training, development, and testing sets.

\paragraph{Symbolic Reasoning Trajectories \& PRM.} Using our proposed method, we generate $1,505$  reasoning trajectories, including $12,448$ atomic steps in total.
For this purpose, we use a random subset of the training sets of FOLIO ($1,003$ questions), FOLIOv2 ($1,000$ questions), and LogicAsker ($300$ questions). 
With these stepwise pseudo-labels, we fine-tune a \textit{Qwen2.5-7B-Instruct} with a binary classification head as a PRM. 

\paragraph{Fine-tuning on Symbolic Reasoning Trajectories.}
We leverage the fine-tuned PRM to select high-quality reasoning trajectories generated by \textit{Llama3.1-8B-Instruct} and \textit{Qwen2.5-7B-Instruct} on all the training questions of FOLIO, FOLIOv2, and LogicAsker.
The PRM predicts a reward (positive or negative) for every step in the reasoning trajectories generated via Symbolic ReAct.
We then select as training data only the trajectories where all steps are labelled as correct and for which the final answer matches the ground-truth.
We use the trajectories to fine-tune \textit{Llama3.1-8B-Instruct} or \textit{Qwen2.5-7B-Instruct} via supervised fine-tuning and direct preference optimization (DPO) \cite{NEURIPS2023_a85b405e}.
To acquire the preference data for DPO, we obtain a real value score for each reasoning trajectory:
1) compute the predictive probabilities for a positive label given each step in the filtered trajectories generated by LLMs.
2) compute the cumulated product of all the probabilities in the trajectory to obtain the score value (i.e., the probability of the entire trajectory being correct).
Equation \ref{dpo_value} shows how to compute the score value $v_{\text{DPO}}$:
\begin{equation}
    v_{\text{DPO}}=\prod_{i=1}^N P(y=1|s_i; \theta_{\text{PRM}}),
    \label{dpo_value}
\end{equation}
where $\theta_{\text{PRM}}$ represents the parameters of the trained PRM model.
For any two trajectories that have a difference larger than a threshold, we include them in the preference tuning data by using the one with the higher value as positive and the lower value as negative.
The threshold is set to $0.25$, which strikes a balance between the data quantity and differentiating the positive and negative examples.
After the filtering, we collect a total of 15K reasoning paths (11M tokens) and 21K DPO pairs (44M tokens).

\section{Experimental Setup}

\subsection{Base LLMs}
We include the following open and frontier LLMs in our experiments: (1) \textit{Llama3.1-8B-Instruct} \cite{grattafiori2024llama3herdmodels}; (2) \textit{Qwen2.5-7B-Instruct} \cite{qwen2.5}; (3) \textit{DeepSeek-V3} \cite{deepseekai2025deepseekv3technicalreport}; (4) \textit{GPT-4o} \cite{openai2024gpt4ocard}; and \textit{GPT-5} \cite{openai2025gpt5card}.

We use the symbolic trajectory data (Section~\ref{sec:data}) to fine-tune \textit{Llama3.1-8B-Instruct} and \textit{Qwen2.5-7B-Instruct}. We also tested models fine-tuned with DPO to compare how it differs from supervised fine-tuning on our reasoning trajectories.

\subsection{Reasoning Approaches}

We employ the following methods for logical inference with LLMs:
\begin{itemize}
    \item \textbf{ReAct} prompting with one human-written demonstration to simulate planning-based reasoning.
    \item \textbf{Symbolic-CoT} \cite{xu-etal-2024-faithful} consists of multi-stage prompts for formalization, planning, solving, and verification. 
    \item \textbf{process-DPO} \cite{jiao-etal-2024-learning} performs DPO based on PRM-rated reasoning trajectories. The PRM is trained on stepwise pseudo-labels produced via Monte Carlo estimation based on vanilla ReAct.
    
\end{itemize}

\subsection{Evaluation Data}

\paragraph{Logical Reasoning.} We first evaluate all models and reasoning methods in-domain on the FOLIO and LogicAsker test sets.

\paragraph{Out-of-domain generalizability}
FOLIO and LogicAsker both consist of partially formalized expressions which only appear in classroom instead of the real world.
To investigate whether the proposed method offers out-of-domain generalisability, we include more realistic datasets.
We use claim verification datasets, where a model needs to determine whether a statement/claim is true given a set of textual evidence. 
We experiment with the following datasets from \citet{dougrezlewis2025assessingreasoningcapabilitiesllms}:

\begin{itemize}
    \item \textbf{Vitamin C} \cite{schuster-etal-2021-get} is a claim verification benchmark infused with challenging cases where the evidence changes over time. The evidences are collected from Wikipedia revisions that modify an underlying fact. The dataset contains contrastive claim-evidence pairs that are nearly identical in language and content but one supports a given claim while the other does not.
    \item \textbf{Climate-FEVER} \cite{diggelmann2021climatefeverdatasetverificationrealworld} is adapted from FEVER \cite{thorne-etal-2018-fever} to online claims about climate. For each claim, the authors retrieve the top five relevant evidence from Wikipedia and present them in the dataset.
    \item \textbf{PHEMEPlus} \cite{dougrez-lewis-etal-2022-phemeplus}, as an extension of the PHEME benchmark, it contains Twitter conversations along with relevant external evidence pertaining to a rumourous claim. A model needs to identify whether a claim is True, False or Unverified. Verification of most claims here requires complex reasoning.
\end{itemize}

We use their sampled subsets, where each dataset consists of 500 samples.

\subsection{Implementation Details}

We full fine-tune the models for $10$ epochs, saving checkpoints after each epoch.
The test results are obtained from the checkpoints that have the highest accuracy on the development set.
We perform the same process of collecting Monte Carlo completions, train a process reward model, select trajectories, and fine-tune for process-DPO baseline, which is based on vanilla ReAct without any guidance of reasoning style.
We first tested process-DPO and our fine-tuned models with only instructions and no demonstrations.
We then tested process-DPO and our fine-tuned models using vanilla ReAct. 
For the off-the-shelf LLMs, we sample for answers using 3 random seeds and report the average. 
All fine-tuning runs were performed on a single NVIDIA H100 GPU.
More details about hyperparameters and settings are presented in Appendix \ref{sec:appendix1}.

\section{Results}

\subsection{Logical Reasoning Results}

Table \ref{tab:main_exp} shows the results on logical reasoning tasks,  our Symbolic ReAct approach (applied to closed and open-weight models), and the fine-tuned open-weight models using our SymbReAct-trace data.

\paragraph{Adding symbolic guidance in ReAct helps increase accuracy.}
As expected, the frontier DeepSeek-V3, GPT-4o, and GPT-5 exhibit higher accuracy than other methods due to their significantly larger model sizes and extensive training. 
Interestingly, integrating Symbolic ReAct still substantially boosts the performance of all three LLMs. 
Notably, GPT-4o shows a 4\% increase in accuracy on FOLIO and a 5\% increase on LogicAsker. 
GPT-5 increases 4\% on FOLIO and 2\% on LogicAsker.
DeepSeek-V3 shows a 1\% increase on FOLIO and a 7\% increase on LogicAsker. 
Symbolic ReAct also improves the performance of Qwen2.5-7B-Instruct on FOLIO and Llama3.1-8B-Instruct on LogicAsker, outperforming their vanilla ReAct counterparts, which do not provide any guidance for reasoning schema.

\begin{table}[!t]
    \centering
    \resizebox{\columnwidth}{!}{%
    \begin{tabular}{clcc}
      \toprule
       & \textbf{Method}  & \textbf{FOLIO} & \textbf{LogicAsker} \\
       \midrule
        \multirow{3}{*}{\rotatebox[origin=c]{90}{GPT-4o}}& ReAct \cite{yao2023react} & $70.10$  & $74.40$   \\
       & SymbCoT \cite{xu-etal-2024-faithful} & $49.51$ & $54.33$   \\
       & Symbolic ReAct (Ours) & $\textbf{74.02}$  & $\textbf{79.33}$   \\
       \midrule
        \multirow{3}{*}{\rotatebox[origin=c]{90}{GPT-5}}& ReAct \cite{yao2023react} & $81.86$  & $87.00$   \\
       & SymbCoT \cite{xu-etal-2024-faithful} & $79.41$ & $62.67$   \\
       & Symbolic ReAct (Ours) & $\textbf{85.78}$  & $\textbf{89.00}$   \\
       \midrule 
        \multirow{3}{*}{\rotatebox[origin=c]{90}{DS-V3}}& ReAct \cite{yao2023react} & $83.33$  & $77.98$   \\
       & SymbCoT \cite{xu-etal-2024-faithful} & $51.96$ & $62.00$  \\
        & Symbolic ReAct (Ours) & $\textbf{84.80}$  & $\textbf{85.00}$   \\
       \midrule
       \midrule
       \multirow{9}{*}{\rotatebox[origin=c]{90}{Llama3.1-8B-Instruct}} 
       & ReAct \cite{yao2023react} & $\textbf{53.43}$ & $52.33$ \\
       & SymbCoT \cite{xu-etal-2024-faithful} & $\textbf{53.43}$  & $37.33$  \\ 
        & Symbolic ReAct (Ours) & $50.16$  & $\textbf{54.11}$   \\ 
        \cmidrule{2-4}
       & process-DPO \cite{jiao-etal-2024-learning} & $61.27$ & $62.00$ \\
       & process-DPO \cite{jiao-etal-2024-learning} + ReAct & $53.92$ & $55.67$ \\
       & Sym. Trajectory-FT (Ours)  & $\textbf{63.24}$ & \textbf{63.67}  \\
       & Sym. Trajectory-FT-DPO (Ours)  & $61.27$ & $60.67$  \\
       & Sym. Trajectory-FT + ReAct (Ours) & $62.75$ & $60.67$  \\
       & Sym. Trajectory-FT-DPO + ReAct (Ours) & $60.29$ & $61.33$  \\
       \midrule
       \midrule
       \multirow{9}{*}{\rotatebox[origin=c]{90}{Qwen2.5-7B-Instruct}} 
       & ReAct \cite{yao2023react} & $57.84$ & $\textbf{69.56}$  \\       
       & SymbCoT \cite{xu-etal-2024-faithful} & $53.76$ & $48.33$\\
        & Symbolic ReAct (Ours) & \textbf{60.78}  & $66.78$   \\ \cmidrule{2-4}
       & process-DPO \cite{jiao-etal-2024-learning} & $61.27$ & $68.00$  \\
       & process-DPO \cite{jiao-etal-2024-learning} + ReAct & $61.27$ & $57.67$  \\
       & Sym. Trajectory-FT (Ours)  & $\textbf{68.14}$ & $\textbf{74.00}$  \\
       & Sym. Trajectory-FT-DPO (Ours)  & $58.33$ & $71.00$  \\
       & Sym. Trajectory-FT + ReAct (Ours) & $64.71$ & $67.00$  \\
       & Sym. Trajectory-FT-DPO + ReAct (Ours) & $58.82$ & $57.33$  \\
       \bottomrule
    \end{tabular}%
    }
    \caption{Accuracy in logical reasoning tasks.}
    \label{tab:main_exp}
\end{table}

\begin{table*}[t]
    \centering
    \resizebox{\linewidth}{!}{%
    \begin{tabular}{clccc}
      \toprule
        & \textbf{Method}                                         &  \textbf{RECV Vitamin C}        & \textbf{RECV Climate Fever}     & \textbf{RECV PHEMEPlus}         \\
       \midrule
       \multirow{6}{*}{Llama} & process-DPO \cite{jiao-etal-2024-learning}         & $82.00_{0.16}$            & $76.13_{1.04}$& $\bf{73.47}_{0.77}$  \\
       & process-DPO \cite{jiao-etal-2024-learning} + ReAct & $78.73_{1.64}$           & $74.67_{1.20}$& $71.00_{1.88}$           \\
       \cmidrule{2-5}
       & Symbolic Trajectory-FT (Ours)                      & $\bf{86.06}_{0.77}$  & $\bf{78.27}_{1.06}$& $72.00_{0.85}$           \\
       & Symbolic Trajectory-FT-DPO (Ours)                  & $80.53_{0.81}$           & $75.80_{0.99}$& $72.87_{0.68}$           \\
       & Symbolic Trajectory-FT + ReAct (Ours)              & $74.40_{1.47}$           & $74.40_{0.28}$& $67.53_{0.96}$           \\
       & Symbolic Trajectory-FT-DPO + ReAct (Ours)          & $80.87_{1.64}$           & $75.87_{2.29}$& $70.67_{2.38}$           \\
       \midrule
       \midrule
       \multirow{6}{*}{Qwen} & process-DPO \cite{jiao-etal-2024-learning}         & $54.27_{0.77}$           & $37.67_{0.74}$& $26.13_{0.96}$           \\
       & process-DPO \cite{jiao-etal-2024-learning} + ReAct & $76.80_{1.14}$           & $56.20_{0.65}$& $59.67_{1.67}$           \\
       \cmidrule{2-5}
       & Symbolic Trajectory-FT (Ours)                      & $\bf{81.93}_{0.25}$           & $58.07_{0.41}$& $\bf{60.40}_{0.57}$  \\
       & Symbolic Trajectory-FT-DPO (Ours)                  & $51.47_{1.64}$           & $33.27_{0.93}$& $22.80_{0.65}$           \\
       & Symbolic Trajectory-FT  + ReAct (Ours)             & $79.40_{0.33}$           & $55.93_{1.76}$& $58.60_{0.43}$           \\
       & Symbolic Trajectory-FT-DPO + ReAct (Ours)          & $77.00_{1.47}$           & $\bf{58.53}_{0.52}$& $58.20_{1.30}$           \\
       \bottomrule
    \end{tabular}%
    }
    \caption{Out-of-domain results on claim verification tasks on the RECV benchmark by \citet{dougrezlewis2025assessingreasoningcapabilitiesllms}}
    \label{tab:OOD_exp_500}
\end{table*}

\paragraph{Fine-tuning on SymbReAct-trace data improves smaller models.}
Notably, the much smaller Qwen2.5-7B model, when fine-tuned with our SymbReAct-trace data, achieves performance comparable to DeepSeek-V3, approximately 2\% lower on FOLIO and 3\% lower on LogicAsker than Deepseek-V3.
It nearly matches GPT-4o's performance on LogicAsker. 
This demonstrates that smaller models, when equipped with a process reward model to select high-quality reasoning trajectories with effective guidance, are beneficiary, even if those trajectories are also generated by smaller models. It is important to note that the 70B models are exclusively used for generating training data for fine-tuning the process reward models. The process reward model itself remains a relatively small model compared to GPT-4o and DeepSeek-V3. Additionally, all models using our approach are considerably smaller than DeepSeek-V3 and GPT-4o, yet achieve competitive results.

\paragraph{Symbolic-guided trajectories are better than the trajectories without guidance.}
When compared to process-DPO, which allows models to choose any reasoning style without explicit guidance, fine-tuning models trained on our symbolic trajectories achieves higher accuracy on both FOLIO and LogicAsker. Limiting the comparison to DPO-tuned models, those fine-tuned on guided trajectories generally exhibit superior performance compared to their unguided counterparts. Exceptions include Qwen on LogicAsker with ReAct, and both Qwen on FOLIO and Llama on LogicAsker when employing direct prompting.

\paragraph{Symbolic Trajectory-FT outperforms DPO.}
We consistently observe that supervised fine-tuning on SymbReAct-trace outperforms DPO. The sole exception is Llama Symbolic Trajectory-FT-DPO, which exhibits  0.66\% higher accuracy than Llama Symbolic Trajectory-FT on LogicAsker. One potential explanation could be the cumulative products of PRMs' predictive probabilities fail to differential the positive trajectories from the negative ones. Whether PRMs' predictive probabilities are well-calibrated enough for producing informative DPO pairs still require further investigation.

\paragraph{ReAct does not improve our fine-tuned reasoning models.}
Regarding the choice between ReAct and direct prompting, the latter generally yields superior performance in most cases. The only exceptions are the DPO-tuned Llama model on LogicAsker and the DPO-tuned Qwen model on FOLIO, though the performance differences in these instances are minimal. This suggests that explicitly instructing LLMs to follow ReAct when they have already been trained on ReAct trajectories may not lead to optimal performance. Instead, directly posing questions appears to be more effective, aligning with the experimental settings employed by \citet{jiao-etal-2024-learning} in their experiments.

\subsection{Out-of-Domain Results}

Table \ref{tab:OOD_exp_500} presents the out-of-domain results of all methods. 
We directly employ the model checkpoints selected for FOLIO and LogicAsker testing, reporting the higher accuracy of the two. 
Models fine-tuned using our proposed approach demonstrate increased accuracy compared to process-DPO in most instances, with the exception of Llama on PHEMEPlus.
The accuracy on RECV Vitamin C is increased by about 28\% with Qwen and 4\% with Llama.
The improvement RECV Climate Fever is 21\% with Qwen and 2\% with Llama.
On RECV PhemePlus, there is a 34\% gain with Qwen.
This suggests that fine-tuning on trajectories with symbolic reasoning generally leads to improved transfer performance on claim verification datasets. 
Our test results align with the trends observed by \citet{dougrezlewis2025assessingreasoningcapabilitiesllms}, where PHEMEPlus exhibits the lowest accuracy, followed by Climate Fever and then Vitamin C. 
PHEMEPlus presents a more challenging setting, especially as it is collected from Twitter/X where there is more noise.

We observe that the Qwen models fine-tuned with process-DPO on logical reasoning datasets show substantial degradations on these claim verification datasets (i.e., the first row for Qwen).
However, the performance recovers when prompting with ReAct (i.e., the second row for Qwen).
Such a phenomenon does not appear in Llama models and all the models fine-tuned with symbolic trajectories.
The exact reason for this is not clear, but we found an anomaly in the generations of the Qwen models where they sometimes produce ``.imgur.com'' in the outputs for unknown reasons.

\subsection{Qualitative Analysis}

We further analyse the reasoning trace generated by our Symbolic Trajectory-FT. 
Questions do not contain symbols except for punctuation.
Example 1 is a trace produced by using Qwen Symbolic Trajectory-FT on a LogicAsker question:

\begin{tcolorbox}[float=htb, colback=green!5, colframe=green!60!black, title=Example 1]
\scriptsize
    1. We have four premises
    \begin{itemize}
        \item $\forall x_{20} \, (\text{Tea}(x_{20}) \rightarrow \text{Squash}(x_{20}))$
        \item $\exists x_{20} \, (\text{Cook}(x_{20}) \lor \text{Tea}(x_{20}))$
        \item $\forall x_{20} \, (\text{Squash}(x_{20}) \rightarrow \text{Cook}(x_{20}))$
        \item $\forall x_{20} \, (\text{Cook}(x_{20}) \rightarrow \text{Squash}(x_{20}))$
    \end{itemize}

    2. We need to determine if there exists an $x_{20}$ such that $\text{Squash}(x_{20})$.
    
    3. From premise 2 ($\exists x_{20} \, (\text{Cook}(x_{20}) \lor \text{Tea}(x_{20}))$), we know that at least one person either cooks or makes tea.

    4. Let's consider two cases:
    \begin{itemize}
        \item \textbf{Case A:} There exists an $x_{20}$ who cooks.\\
        In this case, from premise 4 ($\text{Cook}(x_{20}) \rightarrow \text{Squash}(x_{20})$), we can conclude that $\text{Squash}(x_{20})$ must be true.
        
        \item \textbf{Case B:} There exists an $x_{20}$ who makes tea.\\
        From premise 1 ($\text{Tea}(x_{20}) \rightarrow \text{Squash}(x_{20})$), we can directly conclude that $\text{Squash}(x_{20})$ must be true.
    \end{itemize}
    
    5. Since in both cases we can infer that there exists an $x_{20}$ for whom $\text{Squash}(x_{20})$ is true, we can conclude that $\exists x_{20} \, (\text{Squash}(x_{20}))$.

    Finish [True]

    \label{our_output_example}
\end{tcolorbox}

Example 1 shows that the model has effectively learnt to first formalise the premises in first-order logic and then perform multi-step logical reasoning to derive the validity of the target statement.
Example 2, on the contrary, shows the Qwen model tuned with process-DPO based on vanilla ReAct generates a different style of response, where all reasoning is conducted solely in natural language.

Although in this case both reasoning processes reach the correct answer, symbolic reasoning trajectories tend to be more transparent and efficient, requiring, on average, a smaller number of reasoning steps and fewer generated tokens.
Overall, these examples demonstrate how symbolic reasoning trajectories can support higher levels of abstraction, leading empirically to generalization improvements.

\begin{tcolorbox}[float=htb, colback=yellow!10, colframe=orange!60!black, title=Example 2]
    \scriptsize

    Reasoning steps:
    
    1. Let's analyze the premises one by one: Premise 1: For all $x_{20}$, if $x_{20}$ makes tea, then $x_{20}$ is playing squash. This means that making tea implies playing squash.
    Premise 2: There is at least one $x_{20}$ for which $x_{20}$ will cook or $x_{20}$ will make tea.  This tells us that there exists at least one $x_{20}$ who either cooks or makes tea.  Premise 3: For all $x_{20}$, if $x_{20}$ plays squash, then $x_{20}$ is cooking.  This means that playing squash implies cooking. Premise 4: For all $x_{20}$, if $x_{20}$ cooks, then $x_{20}$ is playing squash.  This means that cooking implies playing squash.
    
    2. Now let's consider the conclusion we need to verify:  There is at least one $x_{20}$ for which $x_{20}$ will play squash.
    
    3. To prove this, we need to find a way to connect the premises to this conclusion. 
    
    4. From Premise 2, we know there exists an $x_{20}$ who either cooks or makes tea.
    
    5. If $x_{20}$ cooks (from Premise 2), then according to Premises 3 and 4, $x_{20}$ must be playing squash (since cooking implies playing squash).
    
    6. Therefore, if there is an $x_{20}$ who cooks, we can conclude that there is at least one $x_{20}$ who plays squash.
    
    7. If $x_{20}$ makes tea (from Premise 2), then according to Premise 1, $x_{20}$ must be playing squash.
    
    8. Therefore, if there is an $x_{20}$ who makes tea, we can also conclude that there is at least one $x_{20}$ who plays squash.
    
    9. Since Premise 2 guarantees the existence of at least one $x_{20}$ who either cooks or makes tea, we have established that there must be at least one $x_{20}$ who plays squash.
    
    Finish [True]
\end{tcolorbox}

\section{Conclusion}

We introduced symbolic ReAct and a fine-tuning pipeline to enhance LLM reasoning by generating high-quality symbolic reasoning trajectories via Monte Carlo estimation and then fine-tuning. The generated data are presented as \textit{SymbReAct-trace}. This approach directly addresses LLMs' susceptibility to memorization over genuine generalization. Our results confirm the effectiveness of our methods in improving logical reasoning and out-of-domain generalization. This suggests that symbolically-guided process supervision can substantially alleviate memorization, paving the way for more reliable and robust LLM reasoning.

\section*{Limitations}

The PRM trained on automatically generated data cannot guarantee that the steps synthetically generated reasoning trajectories are completely correct.
The Monte Carlo estimation relies on the completion of LLMs to estimate the quality, which can make mistakes if the later stage of the reasoning chain is challenging.
That might make the PRM underestimate the correctness of the previous step, leading to incorrect training signals. Despite our results showing that the proposed symbolic reasoning trajectories can achieve a better trade-off between scalability and accuracy compared to natural language trajectories,
more reliable automatic evaluation metrics for the reasoning steps are needed to guarantee formal correctness. Future work might explore methods to efficiently integrate external symbolic solvers to provide additional signals to the PRM \cite{leang2025theorem}.


\section*{Acknowledgments}
This work was supported by the EPSRC [grant number EP/Y009800/1], through funding from Responsible AI UK (KP0016) as a Keystone project.
We acknowledge IT Services at the University of Sheffield, Oxford Advanced Research Computing, and Bristol Centre for Supercomputing for the provision of HPC services.

We thank the anonymous reviewers for their helpful comments. The writing of this paper received minor language-polishing suggestions from Gemini.
In addition, parts of our experimental code were drafted or refactored with assistance from GitHub Copilot; all final implementations were manually reviewed and verified by the authors.

\bibliography{anthology,custom}
\clearpage

\appendix

\section{Additional Details}
\label{sec:appendix1}

Table \ref{tab:hyper} shows the hyperparameters in our model training process.
\begin{table}[ht]
    \centering
    \begin{tabular}{lc}
    \toprule 
    Description & Value \\
    \midrule
       Learning Rate  & $5e-7$ \\
       max\_epoch & $10$ \\
        Batch Size & $8$ \\
        max\_seq\_length & $2,048$ \\
        Optimizer & adamw\_torch\_fused \\
    \bottomrule
    \end{tabular}
    \caption{Hyperparameter settings for fine-tuning.}
    \label{tab:hyper}
\end{table}

We implemented the process-DPO (pDPO) as a baseline.
We generate $1,244$ reasoning trajectories, which have $9,908$ steps in total, by prompting \textit{Llama3.3-70B-Instruct} and Qwen2.5-72B-Instruct on a randomly sampled subset of the training sets of FOLIO ($1,003$ questions), FOLIOv2 ($1,000$ questions), and LogicAsker ($300$ questions).
$10$ completions were generated for each intermediate step in the seed trajectory.
The Monte Carlo estimation took about $250$ GPU hours on an AMD Instinct MI300X.
For sampling the trajectories from off-the-shelf models, we used Ollama\footnote{\url{https://ollama.com/}} apart from the API-based models.
We further use about 42 hours for sampling trajectories to fine-tune the reasoning LLMs.

\section{Additional Results}

Table \ref{tab:direct_accuracy} shows the testing results of zero-shot prompting which includes larger LLMs and QwQ.

\begin{table}[ht]
\centering{\small
\begin{tabular}{l c c}
\toprule
\textbf{Model}  & \textbf{FOLIO} & \textbf{LogicAsker} \\
\midrule
llama3.1:8b     & $56.21_{\textcolor{gray}{2.67}}$ & $56.78_{\textcolor{gray}{1.40}}$ \\
qwen2.5:7b      & $65.69_{\textcolor{gray}{1.74}}$ & $72.67_{\textcolor{gray}{1.70}}$ \\
llama3.3:70b    & $66.99_{\textcolor{gray}{1.29}}$ & $\textbf{77.33}_{\textcolor{gray}{0.82}}$ \\
qwen2.5:72b     & $\textbf{72.88}_{\textcolor{gray}{1.01}}$ & $77.00_{\textcolor{gray}{0.98}}$ \\
QwQ             & $71.57_{\textcolor{gray}{0.80}}$ & $73.78_{\textcolor{gray}{0.57}}$ \\
\midrule
GPT-4o          & $72.55$ & $77.18$ \\
DeepSeek-V3     & $80.39$ & $83.73$ \\
\bottomrule
\end{tabular}
\caption{Mean accuracy of zero-shot prompting with simple instruction. The standard deviations are shown as grey subscript for each model.}
\label{tab:direct_accuracy}}
\end{table}

Table \ref{tab:no_filtering} shows the model performance without using a PRM to filter out the low-quality trajectories.

\begin{table}[ht]
\centering{\small
\begin{tabular}{l c c}
\toprule
\textbf{Model}  & \textbf{FOLIO} & \textbf{LogicAsker} \\
\midrule
Llama3.1:8B w/o PRM     & $62.25$ & $63.00$ \\
Llama3.1:8B with PRM     & $63.24$ & $63.67$ \\
Qwen2.5:7B w/o PRM   & $65.20$ & $73.33$ \\
Qwen2.5:7B with PRM    & $68.14$ & $74.00$ \\
\bottomrule
\end{tabular}
\caption{Model accuracy when using trajectories without PRM filtering. All models are instruction-tuned LLMs.}
\label{tab:no_filtering}}
\end{table}

Table \ref{tab:react_accuracy} shows the comparison between vanilla ReAct and symbolic ReAct based on larger LLMs.
The upper part is tested with the vanilla ReAct, while the bottom part is tested with symbolic ReAct.
The symbolic ReAct has higher accuracy in most of the settings except for Llama3.1-8B-Instruct on FOLIO and Qwen2.5-7B-Instruct/\textit{Qwen2.5-72B-Instruct} on LogicAsker.

\begin{table}[ht]
\centering{\small
\begin{tabular}{l c c c}
\toprule
\textbf{Prompt} & \textbf{Model}  & \textbf{FOLIO} & \textbf{LogicAsker} \\
\midrule
\multirow{4}{*}{\rotatebox[origin=c]{90}{Vanilla}} & llama3.1:8b     & $53.43_{\textcolor{gray}{1.83}}$ & $52.33_{\textcolor{gray}{0.82}}$ \\
& Qwen2.5:7b      & $57.84_{\textcolor{gray}{1.74}}$ & $69.56_{\textcolor{gray}{0.57}}$ \\
& Llama3.3:70b   & $70.26_{\textcolor{gray}{1.29}}$ & $77.33_{\textcolor{gray}{0.54}}$ \\
& Qwen2.5:72b     & $73.86_{\textcolor{gray}{0.23}}$ & $79.78_{\textcolor{gray}{2.35}}$ \\
\midrule
\multirow{4}{*}{\rotatebox[origin=c]{90}{Symbolic}} & llama3.1:8b     & $50.16_{\textcolor{gray}{1.16}}$ & $54.11_{\textcolor{gray}{1.03}}$ \\
& Qwen2.5:7b      & $60.78_{\textcolor{gray}{2.62}}$ & $66.78_{\textcolor{gray}{0.68}}$ \\
& Llama3.3:70b   & $73.04_{\textcolor{gray}{1.44}}$  & $\textbf{80.22}_{\textcolor{gray}{0.63}}$ \\
& Qwen2.5:72b     & $\textbf{74.18}_{\textcolor{gray}{1.85}}$ & $78.11_{\textcolor{gray}{0.57}}$ \\
\bottomrule
\end{tabular}
\caption{Mean accuracy of ReAct prompting with simple instruction. The standard deviations are shown as grey subscript for each model. All models are instruction-tuned LLMs.}
\label{tab:react_accuracy}}
\end{table}

\section{Prompt Examples}
\label{sec:appendix3}

In this section, we show an example of the prompt we used for symbolic ReAct.
Table \ref{symbolic_prompt_table} shows the prompt with symbolic instruction and an example of symbolic ReAct format for sampling the trajectories in Section \ref{sec:sampling}.
Table \ref{completion_prompt} shows the prompt for generating Monte Carlo completions based on preceding steps in Section \ref{sec:prm}.

\begin{table*}
  \begin{center}
  \resizebox{\textwidth}{!}{%
  \begin{tabularx}{\textwidth}{|X|}
\hline
\rowcolor{rowcolor} Input to the LLMs  \\ 
  \hline
  \small
Solve a question answering task by having a Thought, then Finish with your answer. Thought can reason about the current situation. Finish [answer] returns the answer and finishes the task. You will be given context that you should use to help you answer the question. Given the facts and a question, try to define predicates and variables if necessary. Then, parse the problem and the question, formulate them into first-order logic formulas. Next, try to infer the statement presented in the question based on the premises. If it is possible to infer the statement, please answer "True". If it is possible to infer the negation of the statement, please answer "False". If the statement cannot be proofed or disproofed, please answer "Uncertain".\\
\small
In your response, please include a reasoning path to show each step of the inference where you need to specify what logic rule is applied to which premises.
The format of the reasoning step should be like the following:\\
\small
Thought: \{A high-level thought to solve the question\}\\
\small
Action: \{The description of an action to realize the thought\}\\
\small
Observation: \{After applying the action, what would you get? Can be a new formula\}\\
\small
It is not necessary to have a new thought for each step, but an action and an observation should be in each step.\\
\small
The grammar of the first-order logic formular is defined as follows:\\
\small
1) logical conjunction of expr1 and expr2: expr1 $\land$ expr2\\
\small
2) logical disjunction of expr1 and expr2: expr1 $\lor$ expr2\\
\small
3) logical exclusive disjunction of expr1 and expr2: expr1 $\oplus$ expr2\\
\small
4) logical negation of expr1: $\neg$expr1\\
\small
5) expr1 implies expr2: expr1 → expr2\\
\small
6) expr1 if and only if expr2: expr1 $\leftrightarrow$ expr2\\
\small
7) logical universal quantification: $\forall$x\\
\small
8) logical existential quantification: $\exists$x\\
\small
=============Example==============\\
\small
Context: All people who regularly drink coffee are dependent on caffeine. People either regularly drink coffee or joke about being addicted to caffeine. No one who jokes about being addicted to caffeine is unaware that caffeine is a drug. Rina is either a student and unaware that caffeine is a drug, or neither a student nor unaware that caffeine is a drug. If Rina is not a person dependent on caffeine and a student, then Rina is either a person dependent on caffeine and a student, or neither a person dependent on caffeine nor a student.\\
\small
Question: Based on the above information, is the following statement true, false, or uncertain? Rina is either a person who jokes about being addicted to caffeine or is unaware that caffeine is a drug.\\
\small
Thought: We need to translate the information into First Order Logic.\\
\small
Action: Define predicates\\
\small
Observation:\\
\small
Dependent(x) ::: x is a person dependent on caffeine.\\
\small
Drinks(x) ::: x regularly drinks coffee.\\
\small
Jokes(x) ::: x jokes about being addicted to caffeine.\\
\small
Unaware(x) ::: x is unaware that caffeine is a drug.\\
\small
Student(x) ::: x is a student.\\
\small
Action: Translate the statements in the context to logic premises\\
\small
Observation:\\
\small
$\forall$x (Drinks(x) → Dependent(x)) ::: All people who regularly drink coffee are dependent on caffeine.\\
\small
$\forall$x (Drinks(x) $\lor$ Jokes(x)) ::: People either regularly drink coffee or joke about being addicted to caffeine.\\
\small
¬$\exists$x (Jokes(x) → Unaware(x)) ::: No one who jokes about being addicted to caffeine is unaware that caffeine is a drug. \\
\small
(Student(rina) $\land$ Unaware(rina)) $\lor$ ¬(Student(rina) $\lor$ Unaware(rina)) ::: Rina is either a student and unaware that caffeine is a drug, or neither a student nor unaware that caffeine is a drug. \\
\small
¬(Dependent(rina) $\land$ Student(rina)) → (Dependent(rina) $\land$ Student(rina)) $\lor$ ¬(Dependent(rina) $\lor$ Student(rina)) ::: If Rina is not a person dependent on caffeine and a student, then Rina is either a person dependent on caffeine and a student, or neither a person dependent on caffeine nor a student.\\
\small

Thought: Infer on the existing formulas to try to proof the statement by applying logic rules\\
\small
Action: Apply the law of quantifier negation on the third formula\\
\small
Observation:\\
\small
$\forall$x ¬(Jokes(x) → Unaware(x)) ::: It is false that all people who jokes about being addicted to caffeine is unaware that caffeine is a drug. \\
\end{tabularx}}
  \end{center}

\end{table*}

\begin{table*}[t!]
  \begin{center}
  \resizebox{\textwidth}{!}{%
  \begin{tabularx}{\textwidth}{|X|}
\small
Action: Apply instantiation on the formulas which have universial quantifiers.\\
\small
Observation:\\
\small
Drinks(rina) → Dependent(rina) ::: If Rina regularly drinks coffee then she is dependent on caffeine.\\
\small
Drinks(rina) $\lor$ Jokes(rina) ::: Rina either regularly drink coffee or jokes about being addicted to caffeine.\\
\small
¬(Jokes(rina) → Unaware(rina)) ::: It is false that if rina jokes about being addicted to caffeine, then she is unaware that caffeine is a drug. \\
\small
Action: Rewrite the implication in ¬(Jokes(rina) → Unaware(rina)) with disjunction\\
\small
Observation: ¬(¬Jokes(rina) $\lor$ Unaware(rina))\\
\small
Action: Apply De Morgan's Laws on ¬(¬Jokes(rina) $\lor$ Unaware(rina))\\
\small
Observation: ¬(¬Jokes(rina)) $\land$ ¬Unaware(rina)\\
\small
Action: Apply double negation on ¬(¬Jokes(rina))\\
\small
Observation: Jokes(rina)\\
\small
Action: Given that Jokes(rina) is true, any disjunction we introduce is also true.\\
\small
Observation: Jokes(rina) $\lor$ Unaware(rina)\\
\small
Action: Finish [True]

\\
  \hline
  \end{tabularx}}
  \end{center}
  \caption{An example of symbolic ReAct prompt.}
  \label{symbolic_prompt_table}
\end{table*}

\begin{table*}[t!]
  \begin{center}
  \resizebox{\textwidth}{!}{%
  \begin{tabularx}{\textwidth}{|X|}
\hline
  \rowcolor{rowcolor} Input to the LLMs  \\ 
\hline
\small
[Symbolic ReAct prompt in Table \ref{symbolic_prompt_table}]\\
\small
[Previous steps in the seed trajectory]\\
\small
I have wrote the first part of the reasoning path. Please continue the reasoning path.

\\
  \hline
  \end{tabularx}}
  \end{center}
  \caption{An example of completion prompt.}
  \label{completion_prompt}
\end{table*}

\section{Output Examples}

Example 3 was generated by Llama symbolic trajectory-FT on the Vitamin C dataset.
Example 4 was generated by Llama symbolic trajectory-FT-DPO on the Climate Fever.

\begin{tcolorbox}[float=htb, colback=green!5, colframe=green!60!black, title=Example 3]
    \scriptsize
    Step 1: Analyze the context provided, which mentions the HBO miniseries ``John Adams'' and a specific event involving Vice President John Adams casting a tiebreaker vote in favor of ratifying the Jay Treaty.
    
    Step 2: Identify the key elements in the context: ``John Adams'' is a miniseries, and it belongs to HBO.
    
    Step 3: Determine the relevance of the information about Vice President John Adams casting a tiebreaker vote to the question at hand. The event described is a specific scene from the miniseries, but it does not directly relate to the question about whether ``John Adams'' is a miniseries that belongs to HBO.
    
    Step 4: Conclude that the information provided confirms that ``John Adams'' is indeed a miniseries, and it is stated that it belongs to HBO.
    
    Step 5: Based on the context and the information provided, the statement ``John Adams is a miniseries that belongs to HBO'' can be inferred as true.
    
    Finish [True]
    
\end{tcolorbox}

\begin{tcolorbox}[float=htb, colback=yellow!10, colframe=orange!60!black, title=Example 4]
    \scriptsize
    Thought: Determine what the Review concluded about the CRU's work and actions.
    
    Action: Identify the key points mentioned in the context about the Review's findings.
    
    Observation: The Review found that the CRU's work had been ``carried out with integrity'' and had used ``fair and satisfactory'' methods. (From premise 2)
    
    Thought: Analyze the Review's findings in relation to the CRU's actions and integrity.
    
    Action: Look for any information that indicates whether the Review concluded that the CRU's actions were normal and did not threaten the integrity of peer review.
    
    Observation: The Review would also scrutinise the CRU's policies and practices, but it found no evidence of scientific malpractice. (From premise 3 and 5)
    
    Thought: Determine if the Review's findings imply that the CRU's actions were normal and did not threaten the integrity of peer review.
    
    Action: Apply the definition of ``normal'' and ``integrity'' to the Review's findings.
    
    Observation: Since the Review found that the CRU's work was carried out with integrity and used fair and satisfactory methods, and there was no evidence of scientific malpractice, it can be inferred that the CRU's actions were normal and did not threaten the integrity of peer review.
    
    Applying the definition of ``normal'' and ``integrity'' to the Review's findings, we can use the following logical rule: 
    
    If A (work is carried out with integrity) and B (no evidence of scientific malpractice), then it can be inferred that C (actions are normal and do not threaten the integrity of peer review).
    
    Therefore, the statement is: True.
    
    Answer: A) True
    
\end{tcolorbox}

\end{document}